# An improved chromosome formulation for genetic algorithms applied to variable selection with the inclusion of interaction terms

**Chee Chun Gan**  cg8pa@viriginia.edu
*Department of Systems and Industrial Engineering*
*University of Virginia*

**Gerard Learmonth**  jl5c@virginia.edu
*Center for Leadership Simulation and Gaming*
*Center for Large-Scale Computational Modelling*
*Frank Batten School of Leadership and Public Policy*
*University of Virginia*

**Abstract**

Genetic algorithms are a well-known method for tackling the problem of variable selection. As they are non-parametric and can use a large variety of fitness functions, they are well-suited as a variable selection wrapper that can be applied to many different models. In almost all cases, the chromosome formulation used in these genetic algorithms consists of a binary vector of length n for n potential variables indicating the presence or absence of the corresponding variables.

While the aforementioned chromosome formulation has exhibited good performance for relatively small n, there are potential problems when the size of n grows very large, especially when interaction terms are considered. We introduce a modification to the standard chromosome formulation that allows for better scalability and model sparsity when interaction terms are included in the predictor search space. Experimental results show that the indexed chromosome formulation demonstrates improved computational efficiency and sparsity on high-dimensional datasets with interaction terms compared to the standard chromosome formulation.

## 1. Introduction

Variable selection is an integral part of building and refining predictive models. With the recent trend of larger and larger volumes of data becoming available to modelers, automated variable selection procedures are gaining in importance due to the lack of scalability of traditional methods involving modeler judgment and visual analytics when hundreds or thousands of predictors are being considered.

Genetic algorithms (GAs), first pioneered by John Holland in 1975 [1], are an evolutionary heuristic algorithm that have been commonly applied to the problem of variable selection. GAs are based on the evolutionary principles of natural selection and genetic mutation and crossover in order to iteratively optimize a population of candidates using a predefined fitness function. GAs are non-parametric and do not require any assumptions regarding the underlying data other than those necessary for evaluation of the fitness function. As the GA selection process is merely a wrapper, choosing the appropriate fitness function allows the GA to be applied to a large variety of different models.

GAs have been applied to the problem of variable selection in many cases. Vafai and De Jong [2] use a genetic algorithm as a "front end" to rule induction systems for classification problems. Shahamat and Pouyan [3] use principal components analysis, linear discriminant analysis and a genetic algorithm to perform variable selection for a Euclidean-distance based classifier for schizophrenia patients. Bhanu and Lin [4] use a GA as part of feature selection for an automatic target detection system in SAR images.

In almost all applications of GAs to variable selection, the standard GA chromosome formulation consists of a vector of n binary bits, where n is the total number of potential predictors. A chromosome is a vector that contains information about the key parameters in a candidate solution. A value of 0 at vector index i would indicate that the ith variable is not included, while conversely a value of 1 would indicate that the ith variable is included in the candidate solution. For example, Figure 1 below shows a sample chromosome for a model with 6 potential variables. The sample chromosome represents a model with the $2^{nd}$, $3^{rd}$, and $6^{th}$ variables included.

Figure 1 : Sample chromosome for main effects variable selection

| 0 | 1 | 1 | 0 | 0 | 1 |

After formulating the chromosome structure, a number of chromosomes are generated to form the initial population. The generation of the initial population can be performed using a variety of methods, the most common being random generation by selecting each bit value in each chromosome according to a random distribution. The population can also be seeded with "good" solutions found through alternative methods in order to reduce the time spent exploring the solution space for viable solutions. Pre-seeding the population also weights the process more towards exploitation rather than exploration.

Once the initial population has been created, the algorithm proceeds to modify the individual chromosomes in succeeding generations via natural selection. In each generation, the performance of each member chromosome is evaluated using a fitness function which can be specified according to the preference of the modeler.

After determining the fitness levels of all members of the population, a selection procedure is then used to choose several parent chromosomes. One common selection method is tournament selection, where candidates are chosen randomly to participate in a "tournament" during which the fitness values of competing chromosomes are compared, with the winner being selected as a parent chromosome. This parallels the biological process of natural selection where more fit individuals in a population have a greater chance of reproducing and passing on their genes to their offspring. Other selection methods include randomly selecting parent chromosomes with increasing probability corresponding to increasing fitness values, or simply ranking the candidate chromosomes and using the top performers as parents.

Once parent chromosomes have been selected, the crossover operation is used to generate offspring, or child chromosomes. Again, there are various forms of crossover operators used with the underlying notion of combining the genes from multiple (usually two) parent chromosomes into a single offspring. The most basic crossover operator is a fixed point crossover, with the crossover point usually being the midpoint of the chromosome. Figure 2 below shows a simple example of a fixed point crossover with two parent chromosomes A and B, with the crossover point being the chromosome midpoint.

Figure 2 : Fixed point crossover

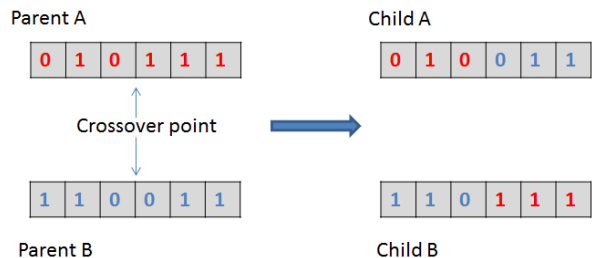

The underlying notion behind the crossover operator is that a high-performing parent chromosome should contain certain elements that contribute to its fitness score. In the case of a variable selection problem, it could be that high performing chromosomes contain a larger ratio of the "correct" variables. By combining the chromosomes of two parents, the crossover operator attempts to generate children which also have a high likelihood of equal or improved performance. The crossover operator can be applied according to a predefined probabilistic parameter setting. For example, a crossover probability of 0.5 would indicate that a pair of parents would have their chromosomes combined half the time. The other half of the time would see both parents being passed on to the next generation without mixing their chromosomes, similar to elitist selection.

The mutation operation (shown in Figure 3) is similar to neighborhood search or hill-climbing methods, where a small change is made to an existing candidate solution in order to explore solutions that are near the original solution in the search space. It is also necessary as a way to introduce novel solutions into the population, as otherwise after several generations the population would lose diversity by consisting only of various recombinations of the original population members. Similar to the crossover operator, the mutation operator is usually applied according to a predefined probabilistic parameter setting.

Figure 3 : Random mutation of single bit

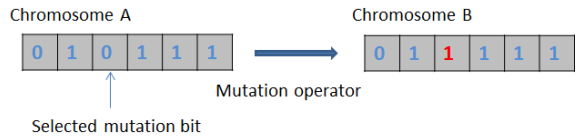

The processes of selection, crossover and mutation taken together form the heart of most GAs. When viewed from the framework of exploration vs exploitation, crossover and mutation serve to explore the solution space in various degrees (crossover provides larger scale changes while mutation can adjust individual bits in the chromosome) while the selection process promotes exploitation of the best currently found solutions by using them as jump off points for exploration. The balance between exploration and exploitation must be adjusted for every application of the GA.

The aforementioned standard chromosome formulation has shown good performance for most variable selection applications. However, the formulation has some shortcomings when applied to very high-dimensional datasets, such as those found when interaction terms are included in the potential search space. We propose an alternative chromosome formulation for GAs applied to variable selection that demonstrates improved performance in terms of run-time, model sparsity and accuracy compared to the standard chromosome formulation.

## 2. Motivation

The standard binary chromosome formulation performs well when the number of potential predictors is relatively small. However, several scalability issues may arise as the number of potential predictors increases. With the increasing interest in analyzing large-scale high-dimensional datasets, the number of potential predictors in some models can easily range from hundreds to thousands. Using the standard chromosome formulation, a variable selection GA would have to keep track of an n-bit vector for each candidate in the population, which can be memory intensive when n is large. Exacerbating the problem is the inclusion of interaction terms, which expand the potential search space combinatorically (for k-way interactions with n predictors, the number of interaction terms is $\binom{n}{k} = \frac{n!}{k!(n-k)!}$ ). For example, by considering only pair-wise interactions the search space of a model with 100 potential main predictors jumps to 5050 possible predictors in total, with a corresponding increase in the amount of memory needed. Furthermore, such a chromosome is also usually very sparse. The vast majority of interaction terms are likely to be uninformative, resulting in a chromosome that is mostly made up of zeros. Thus, in addition to being memory intensive the GA is also memory inefficient.

The "needle in a haystack" structure of searching for interaction terms also poses additional problems to the GA selection procedure using the binary chromosome formulation. When only main effect terms are considered, the distribution of "true" variables can be assumed to be relatively uniform over the length of the chromosome. However, this is no longer true with the inclusion of interaction terms. A very large proportion of variables in the chromosome are uninformative, reducing the probability of the GA selecting a "true" interaction term at each step and therefore reducing the efficiency of the search process via mutation. The large chromosome size also reduces the likelihood of a specific predictor being deleted after entering the chromosome.

### 3. Indexed chromosome formulation

In order to improve scalability, we propose some modifications to the standard chromosome formulation. While only second order interaction terms are examined here, the basic technique for extending the GA framework remains applicable for higher order interactions at the cost of greatly increased computation time. Firstly, a maximum chromosome length l is defined. This allows the modeler to specify an upper bound for model sparsity, as in many instances modelers may not be interested in creating a model with thousands of variables. Secondly, instead of each bit in the chromosome simply being 0-1 to indicate the absence or presence of a variable, each bit now stores the index number of a variable to be included, and 0 if the bit is a "dummy bit". Dummy bits are placeholder bits within the chromosome that reserve space for a potential variable to enter the model. This formulation allows for chromosomes representing models with a differing number of included variables while still allowing chromosome length to be homogenous within the population, which simplifies the crossover operation.

Figure 4 : Chromosome with dummy bits

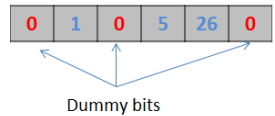

The chromosome in Figure 4 shows a chromosome of length 6 with 3 dummy bits, with variables 1,5 and 26 included in the model. Each new chromosome is initialized with dummy bits in all positions, and the number of initial variables is chosen uniformly between 1 and L (maximum number of variables). Pre-seeded variables can also be utilized instead of random selection. The index positions of these variables are also chosen by sampling without replacement from the available L positions, after which the variables (either randomly chosen or pre-seeded) are then filled into their respective index positions on the chromosome.

The current chromosome formulation can handle an arbitrary number of main effects terms in addition to interaction terms as long as the modeler specifies a maximum number of variables. As the chromosome length is homogenous throughout the population, the aforementioned single point crossover operator can still be applied to the indexed chromosome, with some additional checks to ensure that duplicate variables are removed. However, the mutation operator now has to be separated into two types, a deletion mutation and an addition mutation. The deletion mutation replaces a random non-dummy bit with a value of 0, converting it to a dummy bit and removing the selected variable from the model. The addition mutation replaces a random dummy bit with a randomly selected variable that is currently not included in the model. Both types of mutation occur independently with probabilities $P_a$ and $P_d$ specified by the modeler. Both mutations occur simultaneously with probability $P_a*P_d$, resulting in one variable being switched out for another.

Table 1 compares the probability of adding or deleting a specific variable $x_i$ under the standard chromosome formulation and the indexed formulation, with n total predictors and maximum chromosome length l. Clearly, when $l << n$ and $P_{mutate} = P_a = P_d$ the probability of adding a specific new variable or deleting a specific variable is higher in the indexed formulation, leading to increased variation through mutation.

Table 1 : Comparing addition and deletion probabilities

|  | **Standard** | **Modified** |
|---|---|---|
| **P(adding new variable $x_i$)** | $P_{mutate} * \frac{1}{n}$ | $P_a * \frac{1}{n-l}$ |
| **P(deleting variable $x_i$)** | $P_{mutate} * \frac{1}{n}$ | $P_d * \frac{1}{l}$ |

Recombination using the standard chromosome formulation can also lead to some difficulties in the selection process if interaction terms are included. Intuitively, the easiest way to represent a chromosome with interaction terms using a binary vector is to use the first n bits for n main effect terms, and use the remaining bits for the $\binom{n}{2}$ interaction terms. However, this results in chromosomes with long "tails" that are relatively uninformative compared to the "heads" that contain a large proportion of the predictive power of the model. Performing recombination with standard methods such as fixed single point crossover would result in child chromosomes with reduced variation in fitness as the main effects predictors would tend to always be lumped together in the first half of the chromosome. While this can be mitigated by using more complicated recombination methods as well as reordering the chromosome to spread the interaction terms throughout the length of the chromosome, the indexed chromosome formulation is not as vulnerable to this reduced variation. Firstly, the length of the indexed chromosome is constrained to be much smaller than that of the standard chromosome, which leads to a more even distribution of informative predictors between the "head" and "tail" of the chromosome. Secondly, the predictors in the indexed chromosome are not ordered and are distributed at random along the length of the chromosome depending on the positions of the dummy placeholders, which leads to increased variation during the recombination process.

## 4. Experimental results

Our hypothesis is that the modified GA formulation demonstrates improved performance when applied to large-scale variable selection problems involving interaction terms. To compare the results, both GA formulations were evaluated on a mix of real world and simulated data using a logistic regression model as the underlying fitness function.

In addition to the mutation and fixed point crossover operators outlined previously, we have to ensure that the model obeys strong hierarchy as we are dealing with interaction terms. Each time an interaction term enters the model through either recombination or the addition mutation, a check has to be performed to ensure that the corresponding main effects terms are also included. If not, the missing main effects terms are inserted (either by flipping the corresponding bit index in the standard chromosome or inserting into a random dummy bit position in the modified chromosome). If a main effect term is deleted through the deletion mutation, then all interaction terms that include the aforementioned main effect term are also deleted.

Lastly, in order to prevent selection of models that over-fit the data (a common criticism of using GAs for variable selection), all fitness functions are evaluated using 10-fold cross-validation. The data is partitioned into ten folds, with the models being successively tested on a single fold and trained on the other nine folds. The final fitness is then obtained by averaging the model fitness over all ten test folds. With this process, there is never any overlap between data used for training models, and data used for evaluating the fitness. All experiments were conducted using the statistical package R on an Intel i5 2.7 GHz machine with 8 GB RAM.

### 4.1 UCI Machine Learning Repository datasets

In order to compare the efficacy of both GA formulations, our first set of experiments used a pair of datasets obtained from the UCI Machine Learning Repository [5]. In all experiments, both GAs used the same 10 folds for cross-validation and the same initial seeds, as well as the same meta-parameters.

For our first test case we applied both GA formulation methods using a logistic regression model to a wine quality dataset used by Cortez et al [6]. The dataset consists of 4898 observations of 11 physiochemical properties of red and white variants of Portuguese "Vinho Verde" wine. The output is an integer score between 0 and 10 indicating the quality of the wine. For our purposes, we only considered the white variant as the dataset size was larger, and we transformed the ordinal score into a binary indicator of whether a wine was "good" (score >= 7) or "not good" (score < 7), with a corresponding logistic regression model used as a classifier. The inclusion of pair-wise interaction terms resulted in a total predictor space of 11+55 = 66 terms. The Area under the Receiver Operating Curve (AUC) was used as the fitness measure for both GAs.

Both GAs arrived at very similar final predictor sets (shown in Appendix 1). 11 main effects and 26 interaction terms were selected by both models, while the interaction terms which did not overlap were all non-significant. The AUC for both models was 0.8397, and both models had 6 significant main effects and 20 significant interaction terms (at a 0.05 level), demonstrating that both formulations are able to converge to the same solution when the predictor set is relatively small. However, Table 2 shows the average run time over 5 runs for the GA using the standard chromosome formulation was 2.58 hours, while the average run time over 5 runs for the GA using the indexed chromosome formulation was 2.01 hours, representing a 22% reduction in run time using the same number of generations. Thus, we can see that the indexed chromosome formulation improves the computational efficiency of the GA (with all other parameters held equal) while still being able to converge to the same solution.

Table 2 : Run time for standard vs indexed chromosome GAs (wine quality dataset)

| Run time (hours) | Run 1 | Run 2 | Run 3 | Run 4 | Run 5 | Mean |
|---|---|---|---|---|---|---|
| **Standard** | 2.35 | 2.76 | 2.91 | 2.56 | 2.34 | 2.58 |
| **Indexed** | 1.98 | 2.13 | 1.89 | 1.65 | 2.42 | 2.01 |

The second test case was applied to a cardiotocography dataset provided by Ayres de Campos et al [7]. The datasets consists of 2126 fetal cardiotocograms with 21 predictors (mix of numeric and binary) with the predictand being fetal state (normal, suspect or pathologic) classified using consensus among three expert obstetricians. For our experiment, we transformed the predictand into a binary classifier (normal and abnormal), removed the "Mean", "Median" and "Max" predictors, and applied both GAs to variable selection for an underlying logistic regression model, using AIC as the fitness function. After including interaction terms, the predictor space comprised of 18+153 = 171 terms. The list of predictors and model results are included in Appendix 2.

Similar to the wine quality dataset, both GA formulations obtained results with comparable AIC (461.17 for the standard chromosome, compared to 464.82 for the indexed chromosome). However, the best solution found by the standard chromosome contained 58 terms, with 18 of those terms being main effects and 40 being interaction terms. The GA using the indexed chromosome returned a slightly sparser model with 53 terms (18 main effects terms and 35 interaction terms). There were 24 interaction terms in common between both models, with the main difference being the inclusion of several MSTV interaction terms and the exclusion of several LB and histogram shape interaction terms in the model returned by the GA with the indexed chromosome.

In terms of run time, Table 3 below shows the same trend as Table 2, with the indexed chromosome formulation (3.02 hours) outperforming the standard chromosome formulation (3.64 hours) by an average of 17% over 5 runs.

Table 3 : Run time for standard vs indexed chromosome GAs (cardiotocography dataset)

| Run time(hours) | Run 1 | Run 2 | Run 3 | Run 4 | Run 5 | Mean |
|---|---|---|---|---|---|---|
| Standard | 3.40 | 4.35 | 4.82 | 2.71 | 2.94 | 3.64 |
| Indexed | 3.78 | 2.64 | 3.21 | 2.81 | 2.68 | 3.02 |

### 4.2 Simulated data

The second set of experiments involved using simulated data to compare the performance of the standard GA against the modified GA in variable selection. We used simulated data in order to gain a clearer view of the performance of both GAs when the true predictor set is known for datasets of increasing dimension.

Logistic regression was used again as the underlying model for both GAs. Experiments were run with a population size of 30 for datasets with 5, 20, 30, 40 and 50 main effects predictors. In each experiment, 1000 samples were taken for each predictor from a N(0,1) distribution. A subset s of all predictors (main effects plus pair-wise interactions) is chosen and the predictand y obtained by summing over all predictors $x_i$ in s and adding a N(0,0.02) error term e, then applying a threshold of 2 as shown in the equation below.

$$y = \begin{cases} 0 \text{ if } \sum x_i + e \leq 2 \text{ for } \forall\, x_i \in s \\ 1 \text{ if } \sum x_i + e > 2 \text{ for } \forall\, x_i \in s \end{cases}$$

For the indexed chromosome formulation, a maximum chromosome length of 15, 50, 100, 100 and 100 was defined for datasets with 5, 20, 30, 40 and 50 main effects predictors respectively. The results of the experimental runs are shown below in Table 4, using AIC as the fitness measure for both GAs.

Table 4 : Performance of standard vs indexed chromosome on simulated datasets

|  | Correct terms | Total correct | Model size | AIC | Run time (hours) |
|---|---|---|---|---|---|
| *5 main effects - 15 total predictors* | | | | | |
| Standard | 3 | 3 | 7 | 57 | 0.3179 |
| Indexed | 3 | 3 | 6 | 57 | 0.3309 |
| *20 main effects - 210 total predictors* | | | | | |
| Standard | 19 | 19 | 32 | 66 | 2.87 |
| Indexed | 19 | 19 | 31 | 62 | 0.7774 |
| *30 main effects - 465 total predictors* | | | | | |
| Standard | 27 | 28 | 86 | 174 | 22.959 |
| Indexed | 23 | 28 | 25 | 50 | 0.7574 |
| *40 main effects - 820 total predictors* | | | | | |
| Standard | 31 | 35 | 162 | 326 | 40.728 |
| Indexed | 34 | 35 | 65 | 282 | 0.7850 |
| *50 main effects - 1275 total predictors* | | | | | |
| Standard | N.A. | N.A. | N.A. | N.A. | N.A. |
| Indexed | 33 | 45 | 65 | 587 | 0.7978 |

For the datasets with 5 and 20 main effects, both GAs performed similarly and were able to identify all the correct variables. The standard formulation had a slightly larger model size but had a significantly higher run time for the dataset with 20 main effects terms. For the dataset with 30 main effects terms, the standard formulation managed to pick up 27 out of 28 correct terms, however it utilized 86 terms to do so. The indexed formulation correctly identified 23 out of 28 terms using only 25 terms, which results in a higher AIC. Thus in addition to a much reduced run time, the indexed chromosome is also much sparser as there is a higher chance of selected variables being deleted from the chromosome. The same trend of sparsity and improved run time is shown for the dataset with 40 main effects, except that the indexed chromosome also outperforms the standard chromosome by correctly identifying 34 out of 35 terms (versus 31 out of 35 terms for the standard chromosome). The GA using the standard chromosome formulation was not able to complete execution on the dataset with 50 main effects terms (and 1275 total terms) due to memory issues, while the indexed chromosome formulation was able to correctly identify 33 out of 45 correct terms using 65 terms.

We can also see that the run-time for the indexed chromosome formulation does not increase at the same rate as that for the standard formulation, as the indexed chromosome length is constrained in these experiments to be 100 bits and each experiment is run for 250 generations. However, as the size of the predictor space increases the GA is less and less able to effectively search for good solutions within the allotted number of generations, resulting in decreasing performance in terms of number of correct variables selected unless the number of generations is increased.

## 5. Discussion

Our results show that the indexed chromosome formulation can potentially provide improvements in terms of model sparsity and computational efficiency when applied to large scale variable selection problems where interaction terms are included. By constraining the maximum length of the chromosome and defining the chromosome based on the index positions of the selected variables, the GA is able to more efficiently search through the total predictor space. The indexed chromosome also allows the GA to be more aggressive in pruning uninformative predictors due to the higher chance of predictors being removed compared to the standard chromosome when the chromosome length becomes very large.

The biggest disadvantage of the indexed formulation is the need for the modeler to specify a maximum chromosome size. However, we believe that this is a minor disadvantage as most modelers will have some idea of their desired model size, and the modeler can err on the side of generosity (the probability of adding a variable through mutation does not depend on the number of available spaces). Thus, the modeler can simply increase the maximum chromosome length should the model run out of available space for predictors to enter the chromosome.

# Appendix 1 : Variables selected for wine quality model

| Coefficients: | Standard GA Model | | | | | Indexed GA model | | | | |
|---|---|---|---|---|---|---|---|---|---|---|
| | Estimate | Std. | Error | z-value | Pr(>\|z\|) | Estimate | Std. | Error | z-value | Pr(>\|z\|) |
| (Intercept) | -1.6668 | 0.08981 | -18.559 | < 2.00E-16 | *** | -1.67083 | 0.09005 | -18.555 | < 2.00E-16 | *** |
| **Main Effects** | | | | | | | | | | |
| fixed.acidity | 0.67982 | 0.09752 | 6.971 | 3.15E-12 | *** | 0.6499 | 0.10132 | 6.414 | 1.42E-10 | *** |
| volatile.acidity | -0.81268 | 0.0889 | -9.141 | < 2.00E-16 | *** | -0.79892 | 0.08744 | -9.136 | < 2.00E-16 | *** |
| citric.acid | -0.04427 | 0.06425 | -0.689 | 0.490799 | | -0.04128 | 0.06409 | -0.644 | 0.519561 | |
| residual.sugar | 1.45423 | 0.22534 | 6.453 | 1.09E-10 | *** | 1.44872 | 0.22514 | 6.435 | 1.24E-10 | *** |
| chlorides | -0.57217 | 0.12697 | -4.507 | 6.59E-06 | *** | -0.54172 | 0.12582 | -4.305 | 1.67E-05 | *** |
| free.sulfur.dioxide | 0.15545 | 0.08362 | 1.859 | 0.063014 | . | 0.15195 | 0.08292 | 1.833 | 0.066859 | . |
| total.sulfur.dioxide | 0.05197 | 0.0873 | 0.595 | 0.551654 | | 0.02991 | 0.08568 | 0.349 | 0.727068 | |
| density | -1.81098 | 0.34841 | -5.198 | 2.02E-07 | *** | -1.79446 | 0.34906 | -5.141 | 2.74E-07 | *** |
| pH | 0.5982 | 0.09179 | 6.517 | 7.18E-11 | *** | 0.60901 | 0.09073 | 6.712 | 1.92E-11 | *** |
| sulphates | 0.1211 | 0.06538 | 1.852 | 0.064 | . | 0.10962 | 0.06085 | 1.801 | 0.071633 | . |
| alcohol | 0.23499 | 0.17737 | 1.325 | 0.185216 | | 0.21345 | 0.17363 | 1.229 | 0.218951 | |
| **Interaction Effects** | | | | | | | | | | |
| fixed.acidity:volatile.acidity | 0.21069 | 0.07157 | 2.944 | 0.003243 | ** | 0.20041 | 0.07164 | 2.798 | 0.005149 | ** |
| fixed.acidity:citric.acid | -0.19669 | 0.06167 | -3.189 | 0.001426 | ** | -0.19401 | 0.06181 | -3.139 | 0.001697 | ** |
| fixed.acidity:chlorides | | | | | | -0.13 | 0.12305 | -1.056 | 0.290763 | |
| fixed.acidity:free.sulfur.dioxide | 0.17756 | 0.07694 | 2.308 | 0.021012 | * | 0.17668 | 0.07684 | 2.299 | 0.021484 | * |
| fixed.acidity:total.sulfur.dioxide | -0.2311 | 0.08823 | -2.619 | 0.008811 | ** | -0.20454 | 0.08354 | -2.449 | 0.014342 | * |
| fixed.acidity:pH | 0.11857 | 0.0493 | 2.405 | 0.016178 | * | 0.12464 | 0.04944 | 2.521 | 0.011708 | * |
| fixed.acidity:alcohol | -0.21819 | 0.06997 | -3.118 | 0.00182 | ** | -0.2229 | 0.07219 | -3.088 | 0.002016 | ** |
| volatile.acidity:chlorides | -0.52205 | 0.14052 | -3.715 | 0.000203 | *** | -0.47086 | 0.13169 | -3.575 | 0.00035 | *** |
| volatile.acidity: free.sulfur.dioxide | 0.0935 | 0.07534 | 1.241 | 0.214542 | | 0.08925 | 0.075 | 1.19 | 0.234059 | |
| volatile.acidity:pH | 0.31132 | 0.07746 | 4.019 | 5.84E-05 | *** | 0.30717 | 0.07634 | 4.024 | 5.73E-05 | *** |
| volatile.acidity:sulphates | 0.01886 | 0.05396 | 0.35 | 0.726636 | | 0.01347 | 0.0527 | 0.256 | 0.798319 | |
| volatile.acidity:alcohol | 0.35966 | 0.06598 | 5.451 | 5.01E-08 | *** | 0.36039 | 0.06647 | 5.422 | 5.89E-08 | *** |
| citric.acid:free.sulfur.dioxide | 0.12606 | 0.09007 | 1.4 | 0.161607 | | 0.12349 | 0.08971 | 1.376 | 0.168672 | |
| citric.acid:total.sulfur.dioxide | -0.05545 | 0.08801 | -0.63 | 0.528643 | | -0.05574 | 0.08812 | -0.633 | 0.527018 | |
| residual.sugar: free.sulfur.dioxide | -0.19477 | 0.07898 | -2.466 | 0.013658 | * | -0.18416 | 0.07909 | -2.328 | 0.019888 | * |
| residual.sugar:density | -0.35654 | 0.10628 | -3.355 | 0.000794 | *** | -0.36653 | 0.10658 | -3.439 | 0.000584 | *** |
| residual.sugar:sulphates | | | | | | -0.02387 | 0.09785 | -0.244 | 0.8073 | |
| residual.sugar:alcohol | -0.35549 | 0.10127 | -3.51 | 0.000448 | *** | -0.34311 | 0.10326 | -3.323 | 0.000891 | *** |
| chlorides:total.sulfur.dioxide | 0.17737 | 0.1261 | 1.407 | 0.159549 | | 0.09529 | 0.12739 | 0.748 | 0.454445 | |
| chlorides:density | | | | | | 0.07887 | 0.14599 | 0.54 | 0.589026 | |
| chlorides:pH | -0.24352 | 0.09814 | -2.481 | 0.013087 | * | -0.24962 | 0.0929 | -2.687 | 0.007214 | ** |
| chlorides:alcohol | 0.06992 | 0.11635 | 0.601 | 0.547846 | | | | | | |
| free.sulfur.dioxide: total.sulfur.dioxide | -0.37597 | 0.07299 | -5.151 | 2.59E-07 | *** | -0.37299 | 0.07297 | -5.111 | 3.20E-07 | *** |
| free.sulfur.dioxide:sulphates | 0.41577 | 0.07035 | 5.91 | 3.42E-09 | *** | 0.42687 | 0.0706 | 6.046 | 1.48E-09 | *** |
| free.sulfur.dioxide:alcohol | 0.21913 | 0.08667 | 2.528 | 0.011458 | * | 0.22526 | 0.08611 | 2.616 | 0.008896 | ** |
| total.sulfur.dioxide:density | 0.36846 | 0.09031 | 4.08 | 4.50E-05 | *** | 0.37808 | 0.09033 | 4.185 | 2.85E-05 | *** |
| total.sulfur.dioxide:pH | -0.03666 | 0.07749 | -0.473 | 0.636132 | | | | | | |
| total.sulfur.dioxide:sulphates | -0.45847 | 0.08257 | -5.552 | 2.82E-08 | *** | -0.47545 | 0.08639 | -5.504 | 3.72E-08 | *** |
| density:pH | -0.42201 | 0.11081 | -3.808 | 0.00014 | *** | -0.45455 | 0.10539 | -4.313 | 1.61E-05 | *** |
| density:sulphates | | | | | | 0.06208 | 0.09731 | 0.638 | 0.523465 | |
| pH:sulphates | 0.08178 | 0.04483 | 1.824 | 0.068099 | . | 0.07502 | 0.04693 | 1.599 | 0.109915 | |
| pH:alcohol | -0.37416 | 0.11307 | -3.309 | 0.000935 | *** | -0.38339 | 0.11278 | -3.399 | 0.000675 | *** |
| sulphates:alcohol | -0.04388 | 0.05723 | -0.767 | 0.443214 | | | | | | |

**Logistic Regression model summary statistics (from R)**

**Standard GA Model**                                                      **Indexed GA model**

| | |
|---|---|
| Null deviance: 4093.7  on 3918  degrees of freedom<br>Residual deviance: 2986.6  on 3878  degrees of freedom<br>AIC: 3068.6<br><br>Number of Fisher Scoring iterations: 7<br><br>AUC from 10-fold cross-validation:          0.8397 | Null deviance: 4093.7  on 3918  degrees of freedom<br>Residual deviance: 2986.1  on 3877  degrees of freedom<br>AIC: 3070.1<br><br>Number of Fisher Scoring iterations: 7<br><br>AUC from 10-fold cross-validation:          0.8394 |

# Appendix 2 : Variables selected for cardiotocography dataset

## Legend

| | |
|---|---|
| LB | Fetal heart rate baseline (beats per minute) |
| AC | # of accelerations per second |
| FM | # of fetal movements per second |
| UC | # of uterine contractions per second |
| DL | # of light decelerations per second |
| DS | # of severe decelerations per second |
| DP | # of prolongued decelerations per second |
| ASTV | percentage of time with abnormal short term variability |
| MSTV | mean value of short term variability |
| ALTV | percentage of time with abnormal long term variability |
| MLTV | mean value of long term variability |
| Width | width of FHR histogram |
| Min | minimum of FHR histogram |
| Max | maximum of FHR histogram |
| Nmax | # of histogram peaks |
| Nzeros | # of histogram zeros |
| Mode | histogram mode |
| Mean | histogram mean |
| Median | histogram median |
| Variance | histogram variance |
| Tendency | histogram tendency |

| `Coefficients:` | Standard GA Model | | | | | Indexed GA model | | | |
|---|---|---|---|---|---|---|---|---|---|
| | Estimate | Std. Error | z-value | Pr(>|z|) | | Estimate | Std. Error | z-value | Pr(>|z|) |
| `(Intercept)` | -1.41E+01 | 7.25E+00 | -1.939 | 0.052477 | . | 2.27132 | 9.249172 | 0.246 | 0.806015 |

## Main Effects

| | | | | | | | | | | |
|---|---|---|---|---|---|---|---|---|---|---|
| LB       | -2.92E-01 | 9.83E-02 | -2.969 | 0.00299  | **  | -0.36923 | 0.082256 | -4.489 | 7.16E-06 | *** |
| AC       | -2.57E+00 | 5.89E-01 | -4.356 | 1.32E-05 | *** | -1.2802  | 0.507243 | -2.524 | 0.011608 | *   |
| FM       | -1.49E+00 | 3.62E-01 | -4.125 | 3.71E-05 | *** | -0.36978 | 0.166751 | -2.218 | 0.026586 | *   |
| UC       | -4.17E+00 | 1.46E+00 | -2.85  | 0.004367 | **  | -3.30324 | 1.368104 | -2.414 | 0.015758 | *   |
| ASTV     | 1.69E-01  | 5.43E-02 | 3.12   | 0.001807 | **  | 0.233666 | 0.044817 | 5.214  | 1.85E-07 | *** |
| MSTV     | -1.70E+01 | 7.94E+00 | -2.144 | 0.032022 | *   | -53.8336 | 13.56986 | -3.967 | 7.27E-05 | *** |
| ALTV     | 2.98E-01  | 1.85E-01 | 1.612  | 0.106966 |     | 0.230478 | 0.200253 | 1.151  | 0.249758 |     |
| MLTV     | -1.56E-01 | 7.95E-02 | -1.958 | 0.050205 | .   | 0.103186 | 0.101784 | 1.014  | 0.310691 |     |
| DL       | -1.07E+00 | 2.98E+00 | -0.359 | 0.719914 |     | -3.40643 | 3.666412 | -0.929 | 0.352842 |     |
| DS       | -1.74E+01 | 8.37E+02 | -0.021 | 0.983451 |     | -0.63322 | 815.1317 | -0.001 | 0.99938  |     |
| DP       | 6.24E+01  | 1.39E+01 | 4.494  | 7.00E-06 | *** | 70.65798 | 18.20987 | 3.88   | 0.000104 | *** |
| Width    | 5.86E-02  | 4.64E-02 | 1.264  | 0.206204 |     | -0.06653 | 0.038178 | -1.742 | 0.081422 | .   |
| Min      | 9.51E-02  | 3.13E-02 | 3.037  | 0.002388 | **  | -0.0616  | 0.046181 | -1.334 | 0.182238 |     |
| Nmax     | -1.29E-01 | 1.44E+00 | -0.089 | 0.928725 |     | 0.051742 | 0.121858 | 0.425  | 0.67112  |     |
| Nzeros   | -1.03E+01 | 5.80E+00 | -1.776 | 0.075791 | .   | 1.748274 | 0.711616 | 2.457  | 0.014019 | *   |
| Mode     | 2.23E-01  | 8.75E-02 | 2.547  | 0.010869 | *   | 0.311774 | 0.092868 | 3.357  | 0.000787 | *** |
| Variance | 9.38E-01  | 2.49E-01 | 3.774  | 0.000161 | *** | 1.057749 | 0.326159 | 3.243  | 0.001183 | **  |
| Tendency | 1.48E+01  | 3.91E+00 | 3.795  | 0.000148 | *** | 13.46981 | 3.519186 | 3.828  | 0.000129 | *** |

## Interaction Effects

| | | | | | | | | | | |
|---|---|---|---|---|---|---|---|---|---|---|
| LB:UC     | 3.55E-02  | 9.12E-03 | 3.891  | 9.97E-05 | *** |          |          |       |          |     |
| LB:ALTV   | 9.94E-03  | 2.99E-03 | 3.326  | 0.000882 | *** | 0.010541 | 0.002784 | 3.787 | 0.000153 | *** |
| LB:DL     | 1.42E-01  | 5.36E-02 | 2.646  | 0.00815  | **  | 0.281607 | 0.073479 | 3.832 | 0.000127 | *** |
| LB:Nmax   | -8.65E-02 | 2.30E-02 | -3.766 | 0.000166 | *** |          |          |       |          |     |
| LB:Nzeros | 1.93E-01  | 7.34E-02 | 2.626  | 0.008634 | **  |          |          |       |          |     |
| AC:FM     | -5.49E-02 | 2.02E-02 | -2.715 | 0.006633 | **  |          |          |       |          |     |
| AC:UC     |           |          |        |          |     | -0.27934 | 0.123881 | -2.255 | 0.024138 | *   |
| AC:ALTV | 4.73E-02 | 1.84E-02 | 2.571 | 0.01014 | * | 0.031915 | 0.018517 | 1.724 | 0.084792 | . |
| AC:DL | 3.80E-01 | 2.94E-01 | 1.294 | 0.195623 | | 0.552007 | 0.28532 | 1.935 | 0.053028 | . |
| AC:Variance | -6.62E-02 | 4.01E-02 | -1.652 | 0.098538 | . | -0.12508 | 0.046733 | -2.677 | 0.007438 | ** |
| FM:UC | 4.15E-02 | 1.85E-02 | 2.239 | 0.02518 | * | | | | | |
| FM:ALTV | 1.92E-02 | 5.43E-03 | 3.534 | 0.000409 | *** | 0.01622 | 0.004267 | 3.802 | 0.000144 | *** |
| FM:DP | 3.55E-01 | 1.01E-01 | 3.525 | 0.000424 | *** | | | | | |
| FM:Min | 4.77E-03 | 1.73E-03 | 2.761 | 0.005763 | ** | 0.002587 | 0.001591 | 1.625 | 0.104086 | |
| FM:Mode | 6.84E-03 | 2.04E-03 | 3.362 | 0.000773 | *** | | | | | |
| FM:Variance | | | | | | 0.006886 | 0.002301 | 2.993 | 0.002767 | ** |
| FM:Tendency | -1.29E-01 | 6.51E-02 | -1.986 | 0.047031 | * | | | | | |
| UC:ASTV | -2.05E-02 | 7.60E-03 | -2.701 | 0.006921 | ** | -0.02087 | 0.008377 | -2.492 | 0.012716 | * |
| UC:MLTV | | | | | | -0.06779 | 0.027779 | -2.44 | 0.014673 | * |
| UC:DL | -3.94E-01 | 1.10E-01 | -3.571 | 0.000355 | *** | -0.46128 | 0.127928 | -3.606 | 0.000311 | *** |
| UC:Nmax | 7.80E-02 | 3.37E-02 | 2.315 | 0.020622 | * | 0.126012 | 0.028201 | 4.468 | 7.88E-06 | *** |
| UC:Nzeros | -6.08E-01 | 2.10E-01 | -2.898 | 0.00375 | ** | -0.37293 | 0.16676 | -2.236 | 0.025329 | * |
| | | | | | | 0.031287 | 0.007502 | 4.171 | 3.04E-05 | *** |
| UC:Variance | 4.22E-02 | 1.25E-02 | 3.389 | 0.0007 | *** | 0.053756 | 0.014495 | 3.709 | 0.000208 | *** |
| ASTV:ALTV | -3.19E-03 | 1.32E-03 | -2.425 | 0.015304 | * | -0.00315 | 0.001297 | -2.429 | 0.015156 | * |
| ASTV:DP | -1.10E-01 | 7.38E-02 | -1.492 | 0.135729 | | -0.39956 | 0.110858 | -3.604 | 0.000313 | *** |
| ASTV:Width | 1.67E-03 | 6.79E-04 | 2.465 | 0.013686 | * | | | | | |
| ASTV:Variance | | | | | | 0.007185 | 0.002363 | 3.041 | 0.002356 | ** |
| MSTV:ALTV | -2.55E-01 | 7.31E-02 | -3.484 | 0.000494 | *** | -0.11953 | 0.067134 | -1.781 | 0.074987 | . |
| MSTV:DL | | | | | | -1.51175 | 0.452684 | -3.34 | 0.000839 | *** |
| MSTV:DP | | | | | | 8.6166 | 2.828549 | 3.046 | 0.002317 | ** |
| MSTV:Width | | | | | | 0.132835 | 0.04088 | 3.249 | 0.001157 | ** |
| MSTV:Min | | | | | | 0.117086 | 0.050963 | 2.297 | 0.021593 | * |
| MSTV:Mode | 1.24E-01 | 5.49E-02 | 2.256 | 0.024089 | * | 0.227535 | 0.075585 | 3.01 | 0.00261 | ** |
| MSTV:Variance | | | | | | -0.05534 | 0.037722 | -1.467 | 0.142401 | |
| ALTV:MLTV | 1.22E-02 | 4.02E-03 | 3.043 | 0.002342 | ** | | | | | |
| ALTV:DL | 5.23E-02 | 2.31E-02 | 2.269 | 0.023288 | * | | | | | |
| ALTV:Mode | -9.87E-03 | 2.46E-03 | -4.006 | 6.19E-05 | *** | -0.00995 | 0.002324 | -4.281 | 1.86E-05 | *** |
| ALTV:Variance | 5.81E-03 | 2.86E-03 | 2.028 | 0.042592 | * | 0.004948 | 0.002736 | 1.808 | 0.070545 | . |
| MLTV:DP | | | | | | -1.15562 | 0.391004 | -2.956 | 0.003121 | ** |
| DL:Mode | -1.28E-01 | 3.63E-02 | -3.529 | 0.000416 | *** | -0.23797 | 0.052472 | -4.535 | 5.76E-06 | *** |
| DP:Mode | -3.98E-01 | 1.18E-01 | -3.372 | 0.000746 | *** | -0.41299 | 0.156687 | -2.636 | 0.008395 | ** |
| Width:Min | -1.12E-03 | 3.04E-04 | -3.677 | 0.000236 | *** | | | | | |
| Min:Variance | 2.39E-03 | 1.34E-03 | 1.784 | 0.074464 | . | | | | | |
| Nmax:Mode | 8.50E-02 | 1.90E-02 | 4.478 | 7.54E-06 | *** | | | | | |
| Nzeros:Mode | -9.77E-02 | 6.14E-02 | -1.592 | 0.111317 | | | | | | |
| Nzeros:Variance | 5.83E-02 | 2.36E-02 | 2.471 | 0.013463 | * | 0.038481 | 0.019453 | 1.978 | 0.047915 | * |
| Nzeros:Tendency | -2.84E+00 | 9.47E-01 | -3.003 | 0.002678 | ** | -1.66648 | 0.667303 | -2.497 | 0.012513 | * |
| Mode:Variance | -9.10E-03 | 1.87E-03 | -4.877 | 1.08E-06 | *** | -0.00994 | 0.002499 | -3.98 | 6.90E-05 | *** |
| Mode:Tendency | -9.85E-02 | 2.67E-02 | -3.691 | 0.000223 | *** | -0.09192 | 0.024191 | -3.8 | 0.000145 | *** |

**Logistic Regression model summary statistics (from R)**

| **Standard GA Model** | **Indexed GA model** |
|---|---|
| Null deviance: 1799.55 on 1700 degrees of freedom<br>Residual deviance: 316.82 on 1647 degrees of freedom<br>AIC: 424.82<br><br>Number of Fisher Scoring iterations: 14<br>AIC from 10-fold cross validation :         461.17 | Null deviance: 1799.6 on 1700 degrees of freedom<br>Residual deviance: 304.5 on 1643 degrees of freedom<br>AIC: 420.5<br><br>Number of Fisher Scoring iterations: 14<br>AIC from 10-fold cross validation :         464.82 |